# Generating Simulations of Motion Events from Verbal Descriptions

**James Pustejovsky**
Computer Science Dept.
Brandeis University
Waltham, MA USA
jamesp@cs.brandeis.edu

**Nikhil Krishnaswamy**
Computer Science Dept.
Brandeis University
Waltham, MA USA
nkrishna@brandeis.edu

## Abstract

In this paper, we describe a computational model for motion events in natural language that maps from linguistic expressions, through a dynamic event interpretation, into three-dimensional temporal simulations in a model. Starting with the model from (Pustejovsky and Moszkowicz, 2011), we analyze motion events using temporally-traced Labelled Transition Systems. We model the distinction between *path*- and *manner*-motion in an operational semantics, and further distinguish different types of manner-of-motion verbs in terms of the mereo-topological relations that hold throughout the process of movement. From these representations, we generate minimal models, which are realized as three-dimensional simulations in software developed with the game engine, *Unity*. The generated simulations act as a conceptual "debugger" for the semantics of different motion verbs: that is, by testing for consistency and informativeness in the model, simulations expose the presuppositions associated with linguistic expressions and their compositions. Because the model generation component is still incomplete, this paper focuses on an implementation which maps directly from linguistic interpretations into the Unity code snippets that create the simulations.

## 1 Introduction

Semantic interpretation requires access to both knowledge about words and how they compose. As the linguistic phenomena associated with lexical semantics have become better understood, several assumptions have emerged across most models of word meaning. These include the following:

(1) a. Lexical meaning involves some sort of "componential analysis", either through predicative primitives or a system of types.
b. The selectional properties of predicators can be explained in terms of these components;
c. An understanding of event semantics and the different role of event participants seems crucial for modeling linguistic utterances.

As a starting point in lexical semantic analysis, a standard methodology in both theoretical and computational linguistics is to identify features in a corpus that differentiate the data in meaningful ways; meaningful in terms of prior theoretical assumptions or in terms of observably differentiated behaviors. Combining these strategies we might, for instance, take a theoretical constraint that we hope to justify through behavioral distinctions in the data. An example of this is the theoretical claim that motion verbs can be meaningfully divided into two classes: *manner*- and *path*-oriented predicates (Talmy, 1985; Jackendoff, 1983; Talmy, 2000). These constructions can be viewed as encoding two aspects of meaning: *how* the movement is happening and *where* it is happening. The former strategy is illustrated in (2a) and the latter in (2b) (where $m$ indicates a manner verb, and $p$ indicates a path verb).

(2) a. The ball rolled$_m$.
b. The ball crossed$_p$ the room.

With both of the verb types, adjunction can make reference to the missing aspect of motion, by introducing a path (as in (3a)) or the manner of movement (in (3b)).

(3) a. The ball rolled$_m$ across the room.
b. The ball crossed$_p$ the room rolling.

Differences in syntactic distribution and grammatical behavior in large datasets, in fact, correlate fairly closely with the theoretical claims made by linguists using small introspective datasets.

The path-manner classification is a case where there are data-derived distinctions that correlate nicely with theoretically inspired predictions. More often than not, however, lexical semantic distinctions are formal stipulations in a linguistic model, that often have no observable correlations to data. For example, an examination of the *manner of movement* class from Levin (1993) illustrates this point. The verbs below are all Levin-class manner of motion verbs:

(4) MANNER OF MOTION VERBS: drive, walk, run, crawl, fly, swim, drag, slide, hop, roll

Assuming the two-way distinction between path and manner predication of motion mentioned above, these verbs do, in fact, tend to pattern according to the latter class in the corpus. Given that they are all manner of motion verbs, however, any data-derived distinctions that emerge within this class will have to be made in terms of additional syntactic or semantic dimensions. While it is most likely possible to differentiate, for example, the verbs *slide* from *roll*, or *walk* from *hop* in the corpus, given enough data, it is important to realize that conceptual and theoretical modeling is often necessary to reveal the factors that semantically distinguish such linguistic expressions, in the first place.

We argue that this problem can be approached with the use of minimal model generation. As Blackburn and Bos (2008) point out, theorem proving (essentially type satisfaction of a verb in one class as opposed to another) provides a "negative handle" on the problem of determining consistency and informativeness for an utterance, while model building provides a "positive handle" on both. For our concerns, simulation construction provides a positive handle on whether two manner of motion processes are distinguished in the model. Further, the simulation must specify *how* they are distinguished, the analogue to informativeness.

In this paper, we argue that traditional lexical modeling can benefit greatly from examining how semantic interpretations are contextually and conceptually grounded. We explore a dynamic interpretation of the lexical semantic model developed in Generative Lexicon Theory (Pustejovsky, 1995; Pustejovsky et al., 2014). Specifically, we are interested in using model building (Blackburn and Bos, 2008; Konrad, 2004; Gardent and Konrad, 2000) and simulation generation (Coyne and Sproat, 2001; Siskind, 2011) to reveal the conceptual presuppositions inherent in natural language expressions. In this paper, we focus our attention on motion verbs, in order to distinguish between manner and path motion verbs, as well as to model mereotopological distinctions within the manner class.

## 2 Situating Motion in Space and Time

The interpretation of motion in language has been one of the most researched areas in linguistics and Artificial Intelligence (Kuipers, 2000; Freksa, 1992; Galton, 2000; Levinson, 2003; Mani and Pustejovsky, 2012). Because of their grammatical and semantic import, linguistic interest in identifying where events happen has focused largely on motion verbs and the role played by paths. Jackendoff (1983), for example, elaborates a semantics for motion verbs incorporating explicit reference to the *path* traversed by the mover, from source to destination (goal) locations. Talmy (1983) develops a similar conceptual template, where the path followed by the *figure* is integral to the conceptualization of the motion against a *ground*. Hence, the path can be identified as the central element in defining the location of the event (Talmy, 2000). Related to this idea, both Zwarts (2005) and Pustejovsky and Moszkowicz (2011) develop mechanisms for dynamically creating the path traversed by a mover in a manner of motion predicate, such as *run* or *drive*. Starting with this approach, the localization of a motion event, therefore, is at least minimally associated with the path created by virtue of the activity.

In addition to capturing the spatial trace of the object in motion, several researchers have pointed out that identifying the shape of the path during motion is also critical for fully interpreting the semantics of movement. Eschenbach et al. (1999) discusses the orientation associated with the trajectory, something they refer to as *oriented curves*. Motivated more by linguistic considerations, Zwarts (2006) introduces the notion of an *event shape*, which is the trajectory associated with an event in space represented by a path. He defines a shape function, which is a partial function assigning unique paths to those events involving motion or extension in physical space. This work suggests that the localization of an event makes reference to orientational as well as configurational factors, a view that is pursued in Pustejovsky (2013b). This forces us to look at the various spatio-temporal regions associated with the event participants, and the interactions between them.

These issues are relevant to our present concerns, because in order to construct a simulation, a motion event must be embedded within an appropriate minimal embedding space. This must sufficiently enclose the event localization, while optionally including room enough for a frame of reference visualization of the event (the viewer's perspective). We return to this issue later in the paper when constructing our simulation from the semantic interpretation associated with motion events.

## 3 Modeling Motion in Language

### 3.1 Theoretical Assumptions

The advantage of adopting a dynamic interpretation of motion is that we can directly distinguish path predication from manner of motion predication in an operational semantics (Miller and Charles, 1991; Miller and Johnson-Laird, 1976) that maps nicely to a simulation environment. Models of processes using updating typically make reference to the notion of a state transition (van Benthem, 1991; Harel, 1984). This is done by distinguishing between formulae, $\phi$, and programs, $\pi$. A formula is interpreted as a classical propositional expression, with assignment of a truth value in a specific model. We will interpret specific models by reference to specific states. A state is a set of propositions with assignments to variables at a specific index. Atomic programs are input/output relations ( $[\![\pi]\!] \subseteq S \times S$ ), and compound programs are constructed from atomic ones following rules of dynamic logic (Harel et al., 2000).

For the present discussion, we represent the dynamics of actions in terms of Labeled Transition Systems (LTSs) (van Benthem, 1991).[1] An LTS consists of a triple, $\langle S, Act, \rightarrow \rangle$, where: $S$ is the set of states; $Act$ is a set of actions; and $\rightarrow$ is a total transition relation: $\rightarrow \subseteq S \times Act \times S$. An action, $\alpha \in Act$, provides the labeling on an arrow, making it explicit what brings about a state-to-state transition. As a shorthand for $(e_1, \alpha, e_2) \in \rightarrow$, we will also use $e_1 \xrightarrow{\alpha} e_2$. If reference to the state content (rather than state name) is required for interpretation purposes (van Benthem et al., 1994), then as shorthand for $(\{\phi\}_{e_1}, \alpha, \{\neg\phi\}_{e_2}) \in \rightarrow$, we use, $\boxed{\phi}_{e_1} \xrightarrow{\alpha} \boxed{\neg\phi}_{e_2}$. Finally, when referring to temporally-indexed states in the model, where $e_i@i$ indicates the state $e_i$ interpreted at time $i$, as shorthand for $(\{\phi\}_{e_1@i}, \alpha, \{\neg\phi\}_{e_2@i+1}) \in \rightarrow$, we will use, $\boxed{\phi}_{e_1}^{i} \xrightarrow{\alpha} \boxed{\neg\phi}_{e_2}^{i+1}$, as described in Pustejovsky (2013).

### 3.2 Distinguishing Path and Manner Motion

We will assume that change of location of an object can be viewed as a special instance of a first-order program, which we will refer to as $\nu$ (Pustejovsky and Moszkowicz, 2011).[2]

(5) $x := y$ ($\nu$-transition, where $loc(z)$ is value being updated)
"$x$ assumes the value given to $y$ in the next state."
$\langle \mathcal{M}, (i, i+1), (u, u[x/u(y)]) \rangle \models x := y$
iff $\langle \mathcal{M}, i, u \rangle \models loc(z) = x \wedge \langle \mathcal{M}, i+1, u[x/u(y)] \rangle \models loc(z) = y$

Given a simple transition, a *process* can be viewed as simply an iteration of $\nu$ (Fernando, 2009). However, as (Pustejovsky, 2013a) points out, since most manner motion verbs in language are actually directed processes, simple decompositions into change-of-location are inadequate. That is, they are guarded transitions where the test is not just non-coreference, but makes reference to values on a scale, $\mathcal{C}$, and ensures that it continues in an order-preserving change through the iterations. When this test references the values on a scale, $\mathcal{C}$, we call this a *directed $\nu$-transition ($\vec{\nu}$)*, e.g., $x \preccurlyeq y$, $x \succcurlyeq y$:

(6) $\vec{\nu} =_{df} e_i \xrightarrow{\mathcal{C}? \; \nu} e_{i+1}$.

(7) $\boxed{loc(z) = x}_{e_0} \xrightarrow{\vec{\nu}} \boxed{loc(z) = y_1}_{e_1} \xrightarrow{\vec{\nu}} \ldots \boxed{loc(z) = y_n}_{e_n}$

This now provides us with our dynamic interpretation of directed manner of motion verbs, such as *slide*, *swim*, *roll*, where we have an iteration of assignments of locations, undistinguished except

---
[1]This is consistent with the approach developed in (Fernando, 2009; Fernando, 2013). This approach to a dynamic interpretation of change in language semantics is also inspired by Steedman (2002).

[2]Cf. Groenendijk and Stokhof (1990) for dynamic updating, and Naumann (2001) for a related analysis.

that the values are order-preserving according to a scalar constraint.

This is quite different from the dynamic interpretation of path predicates. Following (Galton, 2004; Pustejovsky and Moszkowicz, 2011), path predicates such as *arrive* and *leave* make reference to a "distinguished location", not an arbitrary location. For example, *the ball enters the room* is satisified when the distinguished location, $D$, (the room) is successfully tested as the location for the moving object. That is, the location is tested against the current location for an object $((loc(x) \neq D)?)$, and retested until it is satisfied $((loc(x) = D)?)$.

(8) $\overbrace{loc(z) = x}^{(loc(x) \neq D)?}|_{e_0} \xrightarrow{\vec{\nu}} \overbrace{loc(z) = y_1}^{(loc(x) \neq D)?}|_{e_1} \xrightarrow{\vec{\nu}} \ldots$
$\underbrace{loc(z) = y_n}_{(loc(x)=D)?}|_{e_n}$

While beyond the scope of the present discussion, it is worth noting that the model of event structure adopted here for motion verbs fits well with most of the major semantic and syntactic phenomena associated with event classes and Aktionsarten.[3]

### 3.3 Mereotopological Distinctions in Manner

Given the formal distinction between path and manner predicates as described above, let us examine how to differentiate meaning within the manner class. Levin (1993) differentiates this class in terms of argument alternation patterns, and identifies the following verb groupings: ROLL, RUN, EPONYMOUS VEHICLE, WALTZ, ACCOMPANY, and CHASE verbs. While suggestive, these distinctions are only partially useful towards actually teasing apart the semantic dimensions along which we identify the contributing factors of manner.

Mani and Pustejovsky (2012) suggest a different strategy involving the identification of semantic parameters that clearly differentiate verb senses from each other within this class. One parameter exploited quite extensively within the motion class involves the mereotopological contraints that inhere throughout the movement of the object (Randell et al., 1992; Asher and Vieu, 1995; Galton, 2000). Using this parameter, we are able to distinguish several of Levin's classes of manner as well as some novel ones, as described in (9), where a class is defined by the constraints that hold throughout the event (where EC is "externally connected", and DC is "disconnected").

(9) For Figure (F) relative to Ground (G):
  a. EC(F,G), throughout motion:
  b. DC(F,G), throughout motion:
  c. EC(F,G) followed by DC(F,G), throughout motion:
  d. Sub-part(F',F), EC(F',G) followed by DC(F',G), throughout motion:
  e. Containment of F in a Vehicle (V).

For example, consider the semantic distinction between the verbs *slide* and *hop* or *bounce*. When the latter are employed in *induced (directed) motion* constructions (Levin, 1993; Jackendoff, 1996), they take on the meaning of manner of motion verbs. Distinguishing between a sliding and hopping motion involves inspecting the next-state content in the motion n-gram: namely, there is a continuous satisfaction of EC(F,G) throughout the motion for *slide* and a toggling effect (on-off) for the predicates *bounce* and *hop*, as shown in (10).

(10) $\overbrace{loc(z) = x}^{\neg DC(x,G)?}|_{e_0} \xrightarrow{\vec{\nu}} \overbrace{loc(z) = y_1}^{DC(x,G)?}|_{e_1} \xrightarrow{\vec{\nu}}$
$\underbrace{loc(z) = y_2}_{\neg DC(x,G)?}|_{e_2}$

With the surface as the *ground* argument, these verbs are defined in terms of two transitions.[4]

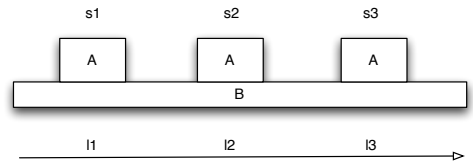

Figure 1: Slide Motion

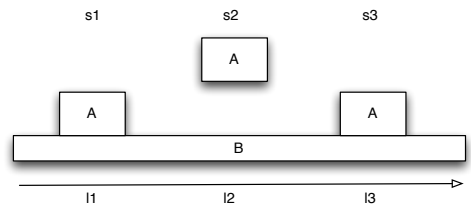

Figure 2: Hop Motion

---

[3]Cf. (Pustejovsky, 2013a) and (Krifka, 1992).

[4]Many natural language predicates require reference to at least three states. These include the semelfactives mentioned above, as well as *blink* and iterative uses of *knock* and *clap* (Vendler, 1967; Dowty, 1979; Rothstein, 2008).

Distinguishing between a sliding motion and a rolling motion is also fairly straightforward. We have the entailments that result from each kind of motion, given a set of initial conditions, as in the following short sentence describing the motion of a ball relative to a floor (the domain for our event simulations).

- *The ball slid.*: At the termination of the action, object `ball` has moved relative to a surface in a manner that is `[+translate]`. That is, the movement is a translocation across absolute space, but other attributes (such as the ball's orientation) do not change.

- *The ball rolled.*: At the termination of the action, object `ball` has moved relative to a surface in a manner that is `[+translate]` *and* `[+rotate]`. Here, the translocation across space is preserved, with the addition of an orientation change.

We can further decompose these features, casting the `[+translate]` in terms of the translation's dimensionality. For both *the ball slid* and *the ball rolled*, it is required that the ball remain in the contact with the relevant surface, thus we can enforce a `[-3-dimensional]` constraint on the `[+translate]` feature. Thus, we arrive at the following differentiating semantic constraints for these verbs: (a) *slide*, `[+translate]`, `[-3-dimensional]`; (b) *roll*, `[+translate]`, `[-3-dimensional]`, `[+rotate]`. This is illustrated below over three states of execution.

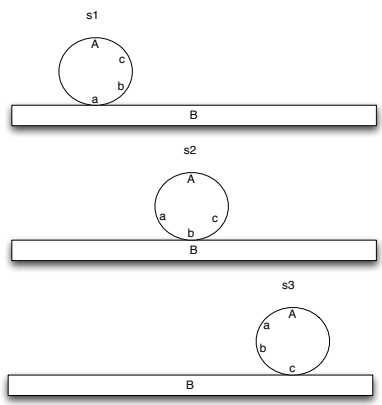

Figure 3: Roll Motion

In our approach to conceptual modeling, we hypothesize that between the members of any pair of motion verbs, there exists at least one distinctive feature of physical motion that distinguishes the two predicates. While this may be too strong, it is helpful in our use of simulations for debugging the lexical semantics of linguistic expressions.[5] In order to quantify the qualitative distinctions between motion predicates and identify the precise primitive components of a motion verb, we build a real-time simulation, within which the individual features of a single motion verb can be defined and isolated in three-dimensional space.

The idea of constructing simulations from linguistic utterances is, of course, not new. There are two groups of researchers who have developed related ideas quite extensively: simulation theorists, working in the philosophy of mind, such as Alvin Goldman and Robert Gordon; and cognitive scientists and linguists, such as Jerry Feldman, Ron Langacker, and Ben Bergen. According to Goldman (1989), simulation provides a process-driven theory of mind and mental attribution, differing from the theory-driven models proposed by Churchland and others (Churchland, 1991). From the cognitive linguistics tradition, simulation semantics has come to denote the mental instantiation of an interpretation of any linguistic utterance (Feldman, 2006; Bergen et al., 2007; Bergen, 2012). While these communities do not seem to reference each other, it is clear from our perspective, that they are both pursuing similar programs, where distinct linguistic utterances correspond to generated models that have differentiated structures and behaviors (Narayanan, 1999; Siskind, 2011; Goldman, 2006).

## 4 Simulations as Minimal Models

The approach to simulation construction introduced in the previous section is inspired by work in minimal model generation (Blackburn and Bos, 2008; Konrad, 2004). Type satisfaction in the compositional process mirrors the theorem proving component, while construction of the specific model helps us distinguish what is inherent in the different manner of motion events. This latter aspect is the "positive handle", (Blackburn and Bos, 2008) which demonstrates the informativeness of a distinction in our simulation.

Simulation software must be able to map a predicate to a known behavior, its arguments to objects in the scene, and then prompt those objects to execute the behavior. A simple input sentence needs to be tagged and parsed and transformed into predicate/argument representation, and from there into a dynamic event structure, as in (Pustejovsky and Moszkowicz, 2011). The event structure is interpreted as the transformation executed over the object or objects in each frame, and then rendered.

---
[5]Obviously, true synonyms in the lexicon would not be distinguishable in a model.

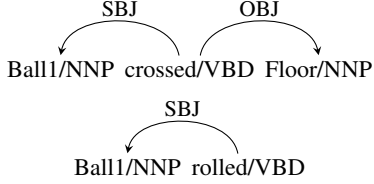

Table 1: Dependency parses for *Ball1 crossed Floor* (top) and *Ball1 rolled* (bottom).

We currently use only proper names to refer to objects in the scene, to simplify model generation, hence *Ball1* and *Floor*. This facilitates easy object identification in this prototype development stage.

Given a tagged and dependency parsed sentence, we can the transform the parse into a predicate formula, using the root of the parse as the predicate, the subject as a singleton first argument, and all objects as an optional stack of subsequent arguments.

| 1. *pred := cross* | 1. *pred := roll* |
| 2. *x := Ball1* | 2. *x := Ball1* |
| 3. *y*.push(*Floor*) | |
| *cross(Ball1,[Floor])* | *roll(Ball1)* |

Table 2: Transformation to predicate formula for *Ball1 crossed Floor* and *Ball1 rolled*.

The resulting predicates are represented in Table 3 as expressions in Dynamic Interval Temporal Logic (DITL) (Pustejovsky and Moszkowicz, 2011), which are equivalent to the LTS expressions used above.

| *cross(Ball1,Floor)* |
|---|
| $loc(Ball1) := y$, $target(Ball1) := z$; $b := y$; $(y := w; y \neq w; d(b,y) < d(b,w)$, $d(b,z) > d(z,w)$, $IN(y,Floor))^+$ |
| *roll(Ball1)* |
| $loc(Ball1) := y$, $rot(Ball1) := z$; $b_{loc} := y$, $b_{rot} := z$; $(y := w; y \neq w; d(b_{loc},y) < d(b_{loc},w)$, $IN(y,Floor))^+$, $(z := v; z \neq v; z\text{-}b_{rot} < v\text{-}b_{rot})^+$ |

Table 3: DITL expressions for *Ball1 crossed Floor* and *Ball1 rolled*.

The DITL expression forms the basis of the coded behavior. The first two initialization steps are coded into the behavior's start function while the the third, Kleene iterated step, is encoded in the behavior's update function.

## 5 Generating Simulations

We use the freely-available game engine, *Unity*, (Goldstone, 2009) to handle all underlying graphics processing, and limited our object library to simple primitive shapes of spheroids, rectangular prisms, and planes. For every instance of an object, the game engine maintains a data structure for the object's virtual representation. Table 4 shows the data structure for `Entity`, the superclass of all movable objects.

| Entity: | |
|---|---|
| position: 3-vector | rotation: 3-vector |
| scale: 3-vector | transform: Matrix |
| collider = { center: 3-vector, min: 3-vector, max: 3-vector, radius: float } | geometry: Mesh |
| currentBehavior: Behavior | |

Table 4: Data structure of motion-capable entities.

The `position` and `scale` of the object are represented as 3-vectors of floating point numbers. The `rotation` is represented as the Euler angles of the object's current rotation, also a 3-vector. This 3-vector is computed as a quaternion for rendering purposes. The `transform` matrix composes the position, scale, and quaternion rotation into the complete transformation applied to the object at any given frame. The `geometry` is a mesh. The points, edges, faces, and texture attributes that comprise the mesh are all immutable at the moment so the mesh type is considered atomic for our purposes. The `collider` contains the coordinates of the center of the object, minimum and maximum extents of the object's boundaries, and radius of the boundaries (for spherical objects).

Behaviors can only be executed over `Entity` instances, so we also provide each one with a `currentBehavior` property, referencing the code to be executed over the object every frame that said behavior is being run. This code performs a transformation over the object at every step, generating a new state in a dynamic model of the event denoted by the a given predicate. Thus, the event[6] is decomposed into frame-by-frame transformations representing the $\nu$-transition from Section 3.2.

We generate example simulations of behaviors in a sample environment, shown in Figure 4, that consists of a sealed four-walled room that contains a number of primitive objects.

---

[6]These events are linguistic events, and not the same as "events" as used in software development or with event handlers.

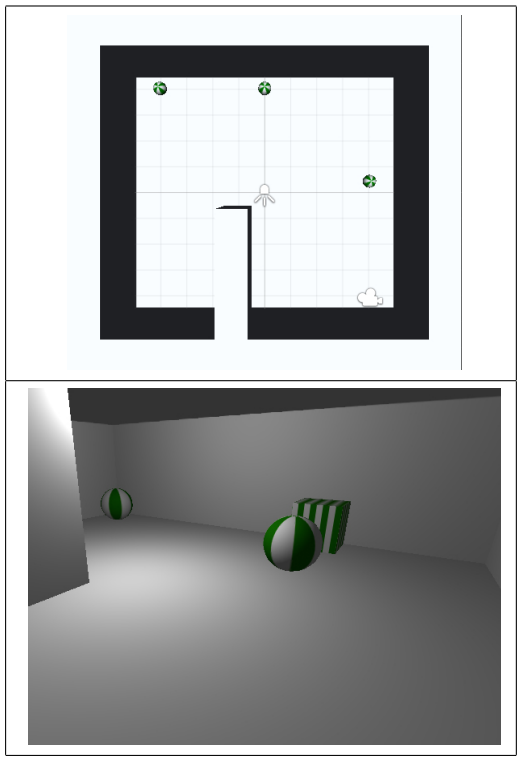

Figure 4: Sample environment in top-down and perspective views.

The behaviors currently coded into our software map directly from DITL to the simulation. The various parts of the DITL formula that describes a given behavior are coded into the behavior's start or update functions in Unity. Below is one such C# code snippet: the per-frame transformation for *roll*.

(11) ```
transform.rotation = new Vector3(
   0.0,0.0,transform.rotation.z+
   (rotSpeed*deltaTime));
transform.position = new Vector3(
   transform.position.x-radius*
   deltaTime,transform.position.y,
   transform.position.z);
```

This "translates" the DITL expression *(y := w; y $\neq$ w; d($b_{loc}$,y) < d($b_{loc}$,w))$^+$, (z := v; z $\neq$ v; z-$b_{rot}$ < v-$b_{rot}$),IN(y,Floor)$^+$* while explicitly calculating the value of the precise differences in location and rotation between each frame or time step. The variables `moveSpeed`, `rotSpeed` and `radius` are given explicit value. `deltaTime` refers to the time elapsed between frames.

Translating a DITL formula into executable code makes evident the differences in minimal verb pairs, such as *the ball (or box) rolled* and *the ball (or box) slid*. When an object rolls, one area on the object must remain in contact with the supporting surface, and that area must be adjacent to the area contacting the surface in the previous time step. When an object slides, *the same* area on the object must contact the supporting surface. Compare the per-frame transformation for *slide* below to the given transformation for *roll*.

(12) ```
transform.position = new Vector3(
   transform.position.x-radius*deltaTime,
   transform.position.y,
   transform.position.z);
```

This maps the DITL expression *(y := w; y $\neq$ w; d($b_{loc}$,y) < d($b_{loc}$,w),IN(y,Floor))$^+$*. Here, the object's location changes along a path leading away from the start location, but does not rotate as in *roll*.

DITL expressions and their coded equivalents can also be composed into new, more specific motions. The *cross* formula from Section 4 can be composed with that for *roll* to describe a "roll across" motion.

In a model, a path verb such as *cross* does not necessarily need an explicit manner of motion specified. In a simulation, the manner needs to be given a value, requiring the composition of the path verb (e.g., *cross*) with one of a certain subsets of manner verbs specifying *how* the object moves relative to the supporting surface. Below are DITL expressions and code implementations for two *cross* predicates, the first a cross motion while sliding, the second a cross motion while rolling.

(13) *loc(Ball1) := y, target(Ball1) := z; b := y; (y := w; y $\neq$ w; d(b,y) < d(b,w), d(b,z) > d(z,w), IN(y,Floor))$^+$*
```
offset = transform.position-
 destination;
offset = Vector3.Normalize(offset);
transform.position = new Vector3(
 transform.position.x-offset.x*
 radius*deltaTime,
 transform.position.y,
 transform.position.z-
 offset.z*radius*deltaTime);
```

At each frame, the distance between the object's current position and its previously computed destination is computed again, and the update moves the object away from its current position *(d(b,y) < d(b,w))* toward the destination *(d(b,z) > d(z,w))*. Since no other manner of motion is specified, the object does not turn or rotate as it moves, but simply "slides."

(14) *loc(Ball1) := y, target(Ball1) := z; b := y; (y := w; y ≠ w; d(b_{loc},y) < d(b_{loc},w), d(b_{loc},z) > d(z,w), (u := v; u ≠ v; u-b_{rot} < v-b_{rot}), IN(y,Floor))$^+$*

```
offset = transform.position-
 destination;
offset = Vector3.Normalize(offset);
transform.rotation = new Vector3(
 0.0,arccos(offset.z)*(360/PI*2),
 transform.rotation.z+
 (rotSpeed*deltaTime));
transform.position = new Vector3(
 transform.position.x-offset.x*
 radius*deltaTime,
 transform.position.y,
 transform.position.z-offset.z*
 radius*deltaTime);
```

Here the update is the same as above, but with the introduction of the rolling motion. In both code snippets, the non-changing value of `transform.position.y` implicitly maps the IN RCC condition in the DITL formulas, and keeps the moving object attached to the floor.

If there exists a behavior corresponding to the predicate (by name) on an entity bearing the name of the predicate's first (subject) argument, the transformation encoded in that behavior is performed over the entity until an end condition specific to the behavior is met. The resulting animated motion depicts the manner of motion denoted by the predicate. Given a predicate of arity greater than 1, the simulator tries to prompt a behavior on the first argument that can be run using parameters of the subsequent arguments.

A *cross* behavior, for example, divides the supporting surface into regions and attempts to move the crossing object from one region to the the opposite region. In figure 5, the bounds of *Floor* completely surround the bounds of *Ball2* (IN(*Ball2*,*Floor*) in RCC8). This configuration makes it possible for the simulation to compute a motion moving the *Ball2* object from one side of the *Floor* to the other.

The left side of figure 5 shows a ball rolling and a box sliding, a depiction of two predicates: *Box1 slid* and *Ball1 rolled*. The right side depicts *Ball2 crossed Floor* (from the rear center to the front center). The starting state of each scene is overlaid semi-transparently while the in-progress state is fully opaque.

## 6 Discussion and Conclusion

In this paper, we describe a model for mapping natural language motion expressions into a 3D simulation environment. Our strategy has been to use minimal simulation generation as a conceptual debugging tool, in order to tease out the semantic differences between linguistic expressions; specifically those between verbs that are members of conventionally homogeneous classes, according to linguistic analysis.

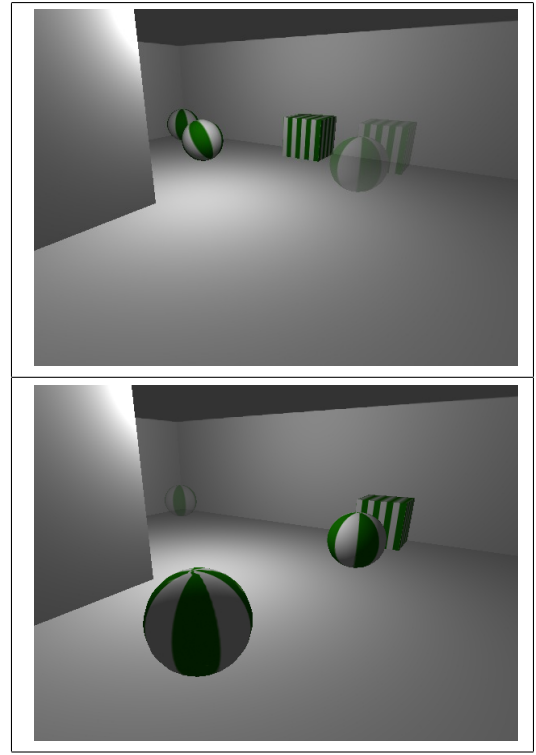

Figure 5: *Roll* and *slide* motions in progress (top), and *cross* motion in progress (bottom).

It should be pointed out that our goal is different from WordsEye (Coyne and Sproat, 2001). While we are interested in using simulation generation to differentiate semantic distinctions in both lexical classes and compositional constructions, the goal behind WordsEye is to provide an enhanced interface to allow non-specialists create 3D scenes without being familiar with special software models for everyday objects and relations. There are obvious synergies between these two goals that can be pursued.

The simulations we create provide *an* interpretation of the given motion predicate over the given entity, but not the *only* interpretation. Just as Coyne et al. (2010) does for static objects in the WordsEye system, we must apply some implicit constraints to our motion predicates to allow them to be visually simulated. For instance, in the *roll* and *slide* examples given in Figure 5, both objects are moving in the same direction–parallel to the back wall of the room object. Had the objects been moving perpendicular to the back wall or in any other direction, as long as they remained in con-tact with the floor at all times, the simulated motion would still be considered a "roll" (if rotating around an axis parallel to the floor), or a "slide" (if not), regardless of what the precise direction of motion is. Minimal pairs in a model have to be compared and contrasted in a discriminative way, and thus in modeling a *slide* predicate versus a *roll* predicate, knowing that the distinction is one of rotation parallel to the surface is enough to distinguish the two predicates in a model.

In a simulation, the discriminative process requires that the two contrasting behaviors look different, and as such, the simulation software must be able to completely render a scene for each frame from behavior start to behavior finish, and so every variable for every object being rendered must have an assigned value, including the position of the object from frame to frame. If these values are left unspecified, the software either fails to compile or throws an exception. Thus, we are forced to arbitrarily choose a direction of motion (as well as direction of rotation, speed of rotation, speed of motion, etc.). As long as all non-changing variables are kept consistent between a minimal pair of behaviors, we can evaluate the quantitative and qualitative differences between the values that do change. As simulations require values to be assigned to variables that can be left unspecified in an ordinary modeling process, simulations expose presuppositions about the semantics of motion verbs and of compositions that would not be necessary in a model alone.

In order to evaluate the appropriateness of a given simulation, we are currently experimenting with a strategy often used in classification and annotation tasks, namely *pairwise similarity judgments* (Rumshisky et al., 2012; Pustejovsky and Rumshisky, 2014). This involves presenting a user with a simple discrimination task that has a reduced cognitive load, comparing the similarity of the example to the target instances. In the present context, a subject is shown a specific simulation resulting from the translation from textual input, through DITL, to the visualization. A set of activity or event descriptions is given, and the subject is then asked to select which best describes the simulation shown; e.g., "Is this a sliding?", "Is this a rolling?". The results of this experiment are presently being evaluated.

The system is currently in the prototype stage and needs to be expanded in three main areas: object library, parsing pipeline, and predicate handling. Our object and behavior libraries are currently limited to geometric primitives and the motions that can be applied over them. While *roll*, *slide*, and *cross* behaviors can be scripted for spheres and cubes and shapes derived from them, a predicate like *walk* cannot be supported on the current infrastructure. Thus, we intend to expand the object library to include more complex inanimate objects (tables, chairs, or other household objects) as well as animate objects. Having an object library containing forms capable of executing greater numbers of predicates will allow us to implement those predicates.

The parsing pipeline described in Section 4 is only partially implemented, with the only completed parts being the latter stages, relating a formulas to a scripted behavior and its arguments. We intend to expand the parsing pipeline to include all the steps described in this paper: taking input as a simple natural language sentence, tagging and parsing it to extract the constituent parts of a predicate/argument representation, and using that output to prompt a behavior in software as a dynamic event structure. More robust parsing will afford us the opportunity to expand the diversity of predicates that the software can handle as well (McDonald and Pustejovsky, 2014). While currently limited to unary and binary predicates, we need to extend the capability to ternary predicates and predicates of greater arity, including the use of adjunct phrases and indirect objects. We are in the process of developing an implementation that uses Boxer (Curran et al., 2007) so that we can create first-order models from the dynamic expressions used here.

## Acknowledgements

We would like to thank David McDonald for comments and discussion. We would also like to thank the reviewers for several substantial suggestions for improvements and clarifications to the paper. All remaining errors are of course the responsibilities of the authors. This work was supported in part by the Department of the Navy, Office of Naval Research under grant N00014-13-1-0228. Any opinions, findings, and conclusions or recommendations expressed in this material are those of the authors and do not necessarily reflect the views of the Office of Naval Research.